\documentclass{ecai}
\usepackage{times}
\usepackage{graphicx}
\usepackage{latexsym}

\usepackage{amsmath}
\usepackage{amssymb}
\usepackage{subcaption}
\usepackage{booktabs}

\begin{document}

\title{Search for Better Students to Learn Distilled Knowledge}

\author{Jindong Gu$^{1,2}$  \and Volker Tresp \institute{The University of Munich, Germany.} $\,^,$\institute{Siemens AG, Corporate Technology, Germany.  \newline email: \{jindong.gu, volker.tresp\}@siemens.com}}

\maketitle
\bibliographystyle{ecai}

\begin{abstract}
Knowledge Distillation, as a model compression technique, has received great attention. The knowledge of a well-performed teacher is distilled to a student with a small architecture. The architecture of the small student is often chosen to be similar to their teacher's, with fewer layers or fewer channels, or both. However, even with the same number of FLOPs or parameters, the students with different architecture can achieve different generalization ability. The configuration of a student architecture requires intensive network architecture engineering. In this work, instead of designing a good student architecture manually, we propose to search for the optimal student automatically. Based on L1-norm optimization, a subgraph from the teacher network topology graph is selected as a student, the goal of which is to minimize the KL-divergence between student's and teacher's outputs. We verify the proposal on CIFAR10 and CIFAR100 datasets. The empirical experiments show that the learned student architecture achieves better performance than ones specified manually. We also visualize and understand the architecture of the found student.
\end{abstract}

\section{Introduction}
Deep learning methods have achieved remarkable performances in many fields, see \cite{lecun2015deep} and the references therein. One of the widely-recognized property of deep neural networks is over-parameterization \cite{Denton2014ExploitingLS}. Such a property requires high computational cost and high memory footprint in forwarding inferences of deep neural networks. The computationally expensive inferences prevent the deploy of deep neural networks in small devices with limited memory size or latency-critical applications such as smartphones and self-driving cars.

Many approaches have been proposed to accelerate the inferences of deep neural networks, such as the parameter quantization approach \cite{Courbariaux2016BinarizedNN,Rastegari2016XNORNetIC,Wu2016QuantizedCN}, Low-rank approximation \cite{Jaderberg2014SpeedingUC,Zhang2015EfficientAA}, Network Pruning \cite{lecun1990optimal,hassibi1993second,han2015learning,Molchanov2017PruningCN,Li2017PruningFF} as well as recently popular Knowledge Distillation \cite{bucilua2006model,hinton2015distilling}.

As a model compression approach, Knowledge Distillation first trains a large teacher network, and then uses its outputs to aid in the training of a smaller student network \cite{Ba2014DoDN,hinton2015distilling}. In this way, the student network can be trained to achieve better performance than if it was trained solely on the training data. The students trained under distillation are closer in performance to their larger teacher. The lower computational cost and memory footprint of the powerful student make its deployment much easier. A large amount of work has been done to improve the distillation process. Many previous works proposed to transfer different (dark) knowledge from the teacher to the student, such as softened label transfer \cite{hinton2015distilling}, feature distillation \cite{Romero2015FitNetsHF}, activation-based and gradient-based spatial attention maps \cite{Zagoruyko2017PayingMA}, derivatives of the loss \cite{Czarnecki2017SobolevTF} and classification boundary \cite{heo2018knowledge}. In this work, we focus on the receiver of the dark knowledge (i.e., the architecture of the student) instead of the dark knowledge itself. Student architectures are often chosen to be smaller architectures by reducing the number of layers or channels of the teacher architecture. How to configure the student architecture so that it can learn distilled knowledge better from their teacher? Manual configuration of network architecture is inefficient since there could be thousands of configuration combinations. In this work, we propose to search for an architecture configuration for the student automatically, instead of designing student architecture manually.

Neural Architecture Search (NAS) has been an active research topic in the machine learning community since \cite{Zoph2017NeuralAS} learns a neural architecture that can achieve competitive performance on the CIFAR-10 using reinforcement learning. Generally, NAS can be identified in three dimensions, namely, search space, search strategy, and performance estimation. The search space is often chain-structured neural networks equipped with modern design elements (e.g., skip connections \cite{he2016deep}). Instead of trying new search spaces exhaustively, we take teacher architecture topology graph as our search space where each channel is taken as a node, weight connections as edges. Therein, we search for a small subgraph as a student architecture. The search strategies in NAS include random search \cite{xie2019exploring}, Bayesian optimization \cite{kandasamy2018neural}, evolutionary methods \cite{Liu2018HierarchicalRF}, reinforcement learning (RL)\cite{Zoph2017NeuralAS}, and gradient-based methods \cite{Liu2019DARTSDA}. The search strategy we use is (sub)gradient-based. We set a gate (a scaling factor) for each node in teacher topology graph, and the scaling factors are regularized by L1-norm to achieve a sparse structure. The scaling factors in L1-norm are updated using its subgradients \footnote{A $g$ is defined as a subgradient of a function $f$, when $f(z) \geq f(x) + g(z-x)$ holds for all $z \in dom f$ given a $x \in dom f$.}. The (sub)gradient-based search strategy is efficient since they do not require costly model performance estimation during the search process.

Another line of research related to our work is network pruning. NAS searches for a small architecture from the search space, which can be reformulated as pruning useless connections from a large network. Our student architecture search can also be seen as pruning teacher architecture. However, our work is different from network pruning. The goal of network pruning is to prune a large model for a smaller model, which achieve comparable performance when it is fine-tuned or re-trained from scratch on the training dataset. With a different goal, we prune teacher architecture for a small student architecture that can learn the distilled knowledge better from the teacher.

The trivial combination of knowledge distillation and network pruning is possible. With no doubt, one can first prune for a small network and train the small network under distillation. However, in the trivial combination, the pruning process is not aware of the late distillation process. Our search process considers the two-step jointly, i.e., pruning for a good student to learn knowledge better from the teacher. In other words, our search process can be seen as distillation-aware network pruning.

Our contributions in this paper are as follows. We define a simple and effective loss function to select student, which defines what a good student architecture should be. We apply L1-normalization on gates specified in the teacher architecture to get a sparse architecture and apply Proximal Gradient Descent to optimize L1 normalization term. Experiments are conducted on the state-of-the-art model and popular datasets to verify our proposal. Furthermore, we provide and analyze the visualization of the found student architecture.

The next section introduces related work. In Section \ref{approach}, we introduce details of our approach, such as the specification of gates in teacher architecture, the loss function, and the optimization method. In experimental Section \ref{approach}, we conduct experiments and analyze experiment results with ablation studies. The last section concludes our work.

\section{Related Work}
\indent \indent \textbf{Knowledge Distillation} 
Rich Caruana and his collaborators demonstrated that the knowledge acquired by a large ensemble of models could be transferred to a single small model \cite{bucilua2006model}. Hinton proposed to train student neural network with the softened labels, which are softened outputs of teacher neural networks \cite{hinton2015distilling}. \cite{Romero2015FitNetsHF} proposed to transfer the knowledge using not the logit layer but earlier feature representations. \cite{Zagoruyko2017PayingMA} transferred the activation-based attentions maps summarised in a forwarding inference and gradient-based attention maps acquired via a backpropagation process to the student. These attention maps can also be applied to understand the decisions made by the underlying neural network. Furthermore, \cite{Czarnecki2017SobolevTF} took derivatives of the loss with respect to inputs as the knowledge to be transferred. Without loss of generality, in this work, we only consider soft labels as the dark knowledge to be transferred, as introduced in the pioneering work \cite{hinton2015distilling}. 

Another closely related work is \cite{Crowley2018MoonshineDW}. The student architecture is often given by empirically shrinking teacher architecture. This work \cite{Crowley2018MoonshineDW} shows that such student is not optimal to learn the dark knowledge. The work replaces the computationally expensive convolutional operations with cheap ones and trains the new model with softened labels extracted from the model before the replacement. The constructed student achieves better performance than the shrunk ones. However, their approach can only handle the teacher with costly operations. The SOTA architecture itself could be already equipped with efficient operations, and it is not clear how to find a cheaper operation for them. We propose an alternative to their method, our method configures the number of layers and channels in each layer of a student architecture automatically, instead of constructing the student architecture with cheap operations manually.

\textbf{Neural Architecture Search} 
\cite{Zoph2017NeuralAS} encoded neural network architectures into numerical sequences and searched for a sequence corresponding to a good network architecture using Reinforcement Learning. The competitive performance of networks found there arouses attention in the machine learning community, although the search process requires thousands of GPU days. Since the neural network architecture space is discrete, the strategies such as random search \cite{xie2019exploring}, reinforcement learning\cite{Zoph2017NeuralAS}, evolutionary methods \cite{Liu2018HierarchicalRF} were applied to tackle the discrete optimization problem. These strategies require large computational power since they have to evaluate hundreds or thousands of intermediate models. To make NAS more efficient, a gradient-based strategy DARTS \cite{Liu2019DARTSDA} were proposed, which transforms a discrete neural network architecture space into a continuous and differentiable form and enables the use of standard gradient-based optimization techniques. In our work, we specify a gate for each channel in the teacher architect. The gate works by multiplying a scaling factor to the activation of the corresponding channel. We apply L1-norm on these scaling factors to force some of them to be zero. L1-norm is not differentiable, but subdifferentiable \footnote{A function $f$ is called subdifferentiable at x if there exists at least one subgradient at all $x \in dom f$.}. A proximal gradient method is applied to update the scaling factors.

\textbf{Network Pruning}
NAS can also be understood from the perspective network pruning where the search space is the large model to be pruned. In a network pruning process, unimportant weights, connections, or neurons can be removed, which often leads to sparse structure \cite{LeCun1989OptimalBD,hassibi1993second,Han2015LearningBW}. The sparse networks can only obtain limited acceleration since most modern hardware and software is optimized for dense matrix multiplication. The work \cite{Li2017PruningFF,huang2018data} prunes the large network channels or connections. The pruning process is also based on L1 normalization. We apply similar techniques to search a student architecture that can learn better from a teacher. Network pruning approaches often remove redundant weights and fine-tuned on the training dataset. \cite{Liu2019RethinkingTV} revisited the value of pruning and showed that it is the pruned architecture, not the inherited weighted contribute to the performance. In our work, we train the student architecture from scratch for a fair comparison.

\section{Student Architecture Search}
\label{approach}
The search space is the topology graph of the teacher model. Each channel in the model is taken as a node, and weights connecting nodes as edges. By removing nodes (channels) and all edges directly connected to those nodes, we can obtain a subgraph, which corresponds to smaller neural network architecture.

Given the teacher topology graph, there are three approaches to achieve a subgraph, namely, non-sturctured pruning, groups sparsity, and structured pruning. The non-structured pruning methods remove the unimportant weights or single neurons. The obtained sparse architecture hardly reduce the inference time on modern hardware. With group sparsity regularization,  the number of neurons can be learned automatically. The parameters there is optimized under both weight decay and group sparse regularization. However, low convergence speed and inferior results resulted from improper optimization technique prevent its applications on modern large scale neural networks. In our search space, channels are taken as individual units. Therefore, we apply a structured pruning method to get student architectures.

\begin{figure}[t]
         \centering
         \includegraphics[width=0.48\textwidth, height=0.2\textwidth]{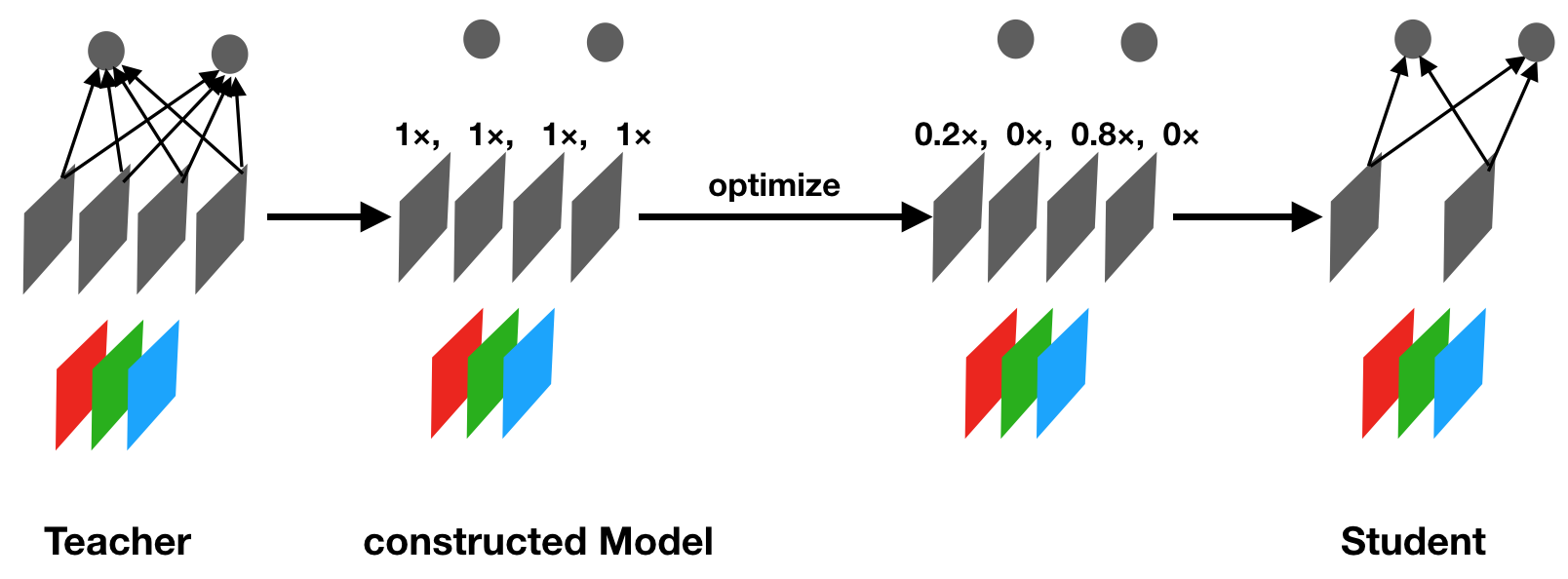}
        \caption{The approach is illustrated on a toy neural network containing an input layer, a single convolution layer, and a fully-connected layer. A new model is constructed on the teacher model by multiplying scaling factors. After an optimization process, the channels with zero scaling factors are removed. The remained small architecture is the selected student architecture.}
        \label{fig:toy_demo}
\end{figure}

\begin{figure}[t]
         \centering
         \includegraphics[width=0.5\textwidth, height=0.16\textwidth]{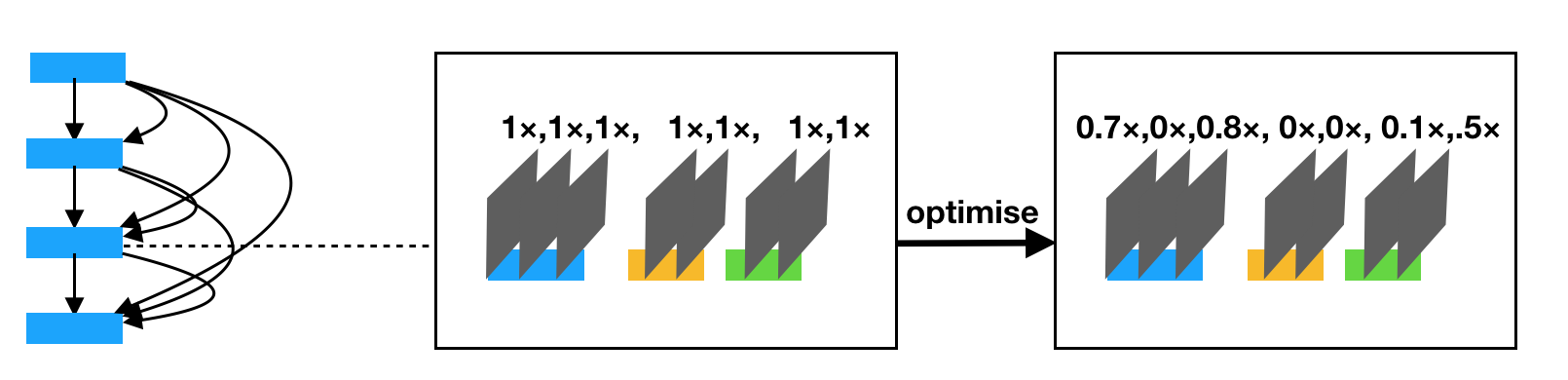}
        \caption{The approach is illustrated on a dense block of DenseNet, which contains four layers. The features of preceding layers are reused in later layers. The dashed line means to zoom in the third layers where the inputs are composed of three parts, i.e., one stem connection and two skip connection. At the end of optimization, the connection is removed if all the gates on its channels are closed.}
        \label{fig:denet_demo}
\end{figure}

More concretely, we specify a gate on each channel by multiplying the activation map of the channel by a scaling factor $g$. At the end of the optimization, the open gate ($g\neq 0$) means the corresponding channel is important to the distillation process, while the closed gate ($g=0$) means the corresponding channel can be removed safely. For a layer with $K$ channels in a teacher neural network, the corresponding $\boldsymbol{g}$ is a K-element vector. The number of channels of the obtained student architecture in this layer is identified by the number of non-zero elements in the vector $\boldsymbol{g}$. A simple demonstration is shown in Figure \ref{fig:toy_demo}.

This method can be easily applied to SOTA neural networks. In this work, we mainly demonstrate our idea on DenseNet \cite{Huang2017DenselyCC}. DenseNet is composed of several stacked Dense blocks. For each layer in a Dense block, the feature-maps of all preceding layers are used as inputs, and its own feature-maps are used as inputs into all subsequent layers. We specify gates for all channels, including the ones from preceding layers. A skip connection between a preceding layer and current layer can be removed if all the gates of the channels in the connection are closed. An illustration on a dense block of DenseNet can be found in Figure \ref{fig:denet_demo}.

\subsection{Distillation-aware Loss function} 
The method to obtain a smaller sub-graph from the teacher topology graph is introduced above. In this subsection, we describe the loss function we optimize to obtain a good student architecture. Since we aim to find a good student, the loss function should be aware of the distillation process.

We first provide background information about the knowledge distillation process. The idea behind knowledge distillation is to train the student network with more information than one-hot ground-truth labels. Given logits of a student network $a_s$  and the one-hot ground truth $y_{gt}$, in classic supervised learning, the loss is usually specified as 
\begin{equation}
L_{CE} = CE(softmax(a_s/\tau), y_{gt}).
\end{equation}
where $CE$ is the cross-entropy loss fucntion and the temperation $\tau$ is set to 1. In knowledge distillation process, the logits of teacher network $a_t$ can offer more information to train the student network. One way to leverage such information is to match the softened outputs of student $softmax(a_s/\tau)$ and teacher $f_t(\mathbf{x_i}) = softmax(a_t/\tau)$ via a KL-divergence loss
\begin{equation}
L_{KD} = KL(softmax(a_s/\tau), softmax(a_t/\tau)).
\end{equation}
where $\tau$ as a hyperparameter is often set bigger than 1 to soften the outputs. The overall loss to train the student network is $L_s = L_{KD} + \lambda L_{CE}$ where the hyperparameter $\lambda$ is often set to a very small value, the second term works by regularizing the training process.

Given an input $\mathbf{x_i}$, the softened output of a teacher model is $f_t(\mathbf{x_i})$, and the softened output of the model constructed by adding gates is $f_s(\mathbf{x_i, w, g})$, i.e., the constructed model in Figure \ref{fig:toy_demo}. The weights and scaling factors therein are updated during the optimization. The loss function we propose is mathematically defined as follows.
\begin{equation}
\small
\min_{\mathbf{w, g}}  \frac{1}{N} \sum^N_{i=1} KL(f_s(\mathbf{x_i, w, g}), f_t(\mathbf{x_i})) + \lambda_1 \|\mathbf{w}\|_2 + \lambda_2 \sum^M_{j=1} \alpha_j \|g_j\|_1
\label{equ:loss}
\end{equation}
where $N$ is the number of training examples, M is the number of gates (channels in whole teacher network), and the hyperparameters $\lambda_1$ and $\lambda_2$ specify the regularization strength for weight decay and L1 normalization respectively. $\alpha_j$ is the weight specified for the channel with the gate $g_j$.

The loss function we propose is composed of three terms. The first term computes the mismatch between softened outputs of the constructed network and that of the teacher network, which corresponding to $L_{KD}$ in knowledge distillation. The selected student architecture can learn distilled knowledge better by matching their softened outputs better. If the first term is replaced by the cross-entropy loss between the outputs of the student network and teacher network, the loss function becomes a target function of network pruning. Since the loss function of network pruning is agnostic to the distillation process, the selected architecture perform worse than the one selected by our distillation-aware loss function.

As a popular regularization technique, weight decay is applied to the weights of the constructed model. The third term is the L1 normalization on all scaling factors corresponding to the specified gates. Since the reduced size can be different when channels in different layers are removed, we weight the scaling factors in L1 normalization term. Intuitively, the gates on channels which contribute more FLOPs in forwarding inferences should be first closed since its removal can save more computational cost. If the saved FLOPs is $F_j$ when the channel with the gate $g_j$ is removed, the weight for the scaling factor $g_j$ is computed as $a_j = \frac{F_j}{\max^M_{k=1} F_k} $.

At the end of the optimization, we remove all the channels with closed gated from the constructed model. The remaining small architecture is taken as the student architecture. Early stopping is applied to obtain an architecture with a certain number of FLOPs.

\begin{figure*}[t]
     \begin{subfigure}{\textwidth}
         \centering
         \includegraphics[width=0.91\textwidth]{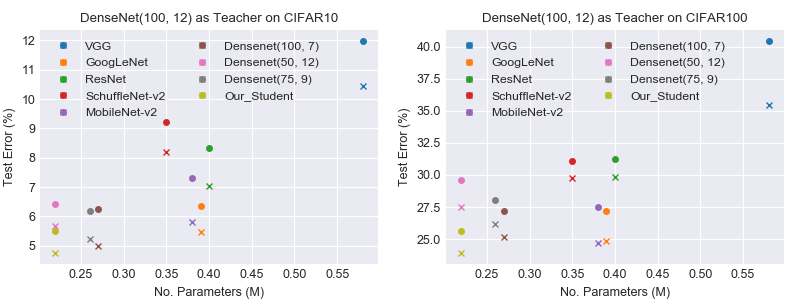}
         \caption{Comparison of students in terms of the number of parameters (the model size).}
         \label{fig:perf_para}
     \end{subfigure}
     
     \begin{subfigure}{\textwidth}
         \centering
         \includegraphics[width=0.91\textwidth]{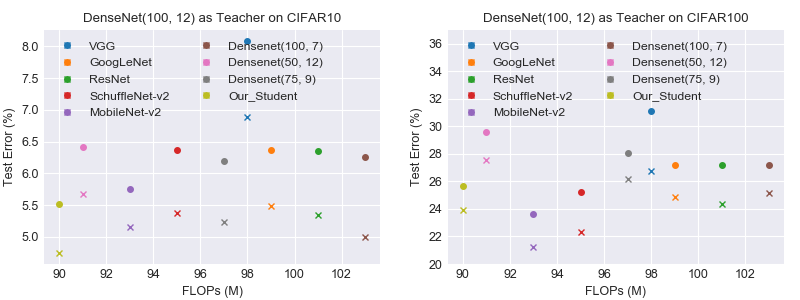}
         \caption{Comparison of students in terms of FLOPs (the cost of forwarding inferences).}
         \label{fig:perf_flops}
     \end{subfigure}
     
        \caption{All student architectures are trained with DenseNet(100, 12) as their teacher. The performance of all students on the two datasets is shown in this figure. In each plot, each color stands for one student architecture. Each color also corresponds to two marks, and the circle mark means the performance trained from scratch while the cross mark describes the performance trained under distillation. Y-axis describes the test error on the corresponding test datasets.}
        \label{fig:perf}
\end{figure*}

\subsection{Optimization of the Loss function} 
In the last subsection, we introduce our distillation-ware loss function. The weights $\mathbf{w}$ and the scaling factors $\mathbf{g}$ are updated to minimize the loss function. The loss function is differentiable to the weights. The weights $\mathbf{w}$ can be updated by Stochastic Gradient Descent (SGD) with momentum or its variants. However, SGD is not applicable to update the scaling factors $\mathbf{g}$ since the loss function is not differentiable to $\mathbf{g}$.

Accelerated Proximal Gradient (APG) \cite{parikh2014proximal} can be applied to update $\mathbf{g}$ since the loss is subdifferentiable to $\mathbf{g}$. However, this optimization method could be computationally expensive for deep neural networks. To overcome this challenge, \cite{huang2018data} reformulates the original APG to avoid redundant forward and backward pass in calculating the gradients. In this work, we also applied this modified version of APG to update scaling factors. We use convienient notation $\ell(\mathbf{g}) = \frac{1}{N} \sum^N_{i=1}  KL(f_s(\mathbf{x_i, w, g}), f_t(\mathbf{x_i}))$. The update rule is as follows:

\begin{equation}
\begin{split}
z(t) &= \mathbf{g}_{(t-1)} -  \eta \cdot \nabla \ell(\mathbf{g}_{(t-1)})  \\
v(t) &= S_{\eta,\gamma}(z(t)) - \mathbf{g}_{(t-1)} + \mu \cdot v(t-1) \\
\mathbf{g}_{(t)} &= S_{\eta,\gamma}(z(t)) + \mu \cdot v(t) \\
\end{split}
\label{equ:rule}
\end{equation}
where $t$ is the number of iterations, $S_{\eta,\gamma}$ is the soft-threshold operator $S_{\alpha}(\mathbf{z})_i = sign(z_i)(|z_i| - \alpha)_+$, $\eta$ means gradient step size and $\mu$ is the momentum. The weights $\mathbf{w}$ and the scaling factors $\mathbf{g}$ are updated jointly on the same training set. At the end of the optimization, parts of scaling factors are forced to be zeros by optimizing L1 normalization with the above rule.

\section{Experiment}
In this section, we first show the performance of the found student on CIFAR10 and CIFAR100 datasets. Then we verify our proposal via ablation studies. We also visualize the found student architecture and provide our understanding.

\begin{table*}[t]
  \caption{Performance of models selected under different loss functions on CIFAR10 and  CIFAR100 (Test Error \%)}
  \label{tab:ablation}
  \centering  
  \begin{tabular}{|l|c|c|c|c|}
  \hline
  Loss & C10 - NOKD &  $\quad$C10 - KD$\quad$ & C100 - NOKD &  $\quad$C100 - KD$\quad$ \\
   \hline
  NP   &  6.31  & 5.48  & 27.81  & 26.35  \\
   \hline
  NP$ _{weighted}$  &  5.98  & 5.18  & 25.97  &  24.22  \\
   \hline
  KD     &  6.55   &  5.54  &  28.49  & 26.94 \\
  \hline
  KD$ _{weighted}$  &  \textbf{5.51}  &  \textbf{4.74}  & \textbf{25.64}   &  \textbf{23.89} \\
  \hline
  \end{tabular}
  \vspace{0.5cm}
\end{table*}

\subsection{Student Search on CIFAR10 and CIFAR100}
We take DenseNet as a teacher model. The teacher model and all student models are trained with the same setting as in \cite{Huang2017DenselyCC}. CIFAR10 and CIFAR100 contain 50,000 training images and 10,000 test images with 32 $\times$ 32 pixels respectively. A standard data augmentation scheme is used for these two datasets. The weight decay is set to 1e-4. All the models are trained for 300 epochs with a batch size of 128. The initial learning rate is set to 0.1, and is divided by 10 at the 150th and the 255th epoch. The moment of 0.9 is used. When students are trained with knowledge distillation, the hyperparameters are $\tau=4, \lambda=0.1$, following \cite{Crowley2018MoonshineDW}. When searching for student architectures, we initialize the constructed model with parameters from the teacher model, set $\lambda_2$ to 1e-3 and update weights $\mathbf{w}$ with the same setting as above. The learning step for updating scaling vectors $\mathbf{g}$ is 0.01 without decay. The hyperparameters in updating rule of Equation \ref{equ:rule} is set the same as in \cite{huang2018data}.

The size of DenseNet is characterized by two indicators, namely the number of layers $L$ and the growth rate of the network $k$. In a dense block with $l$ layers, the number of output channels is $k_0 + k \times (l-1)$ where $k_0$ is the number of channels in the input of the dense block. In our experiments on both CIFAR10 and CIFAR100, the teacher model is DenseNet($L=100, k=12$), noted as DenseNet(100, 12). The number of FLOPs and parameters are $296M$ and $0.8M$ respectively. We specify different students architecture as baselines by manually reducing the number of layers, channels, or both. The specified student architectures are DenseNet(50, 12), DenseNet(100, 7), DenseNet(75, 9). What is more, we also take the SOTA architectures as student architectures, namely, VGG\cite{Simonyan2015VeryDC}, GoogLeNet\cite{Ioffe2015BatchNA}, ResNet\cite{he2016deep}, Shufflenet-v2\cite{Ma2018ShuffleNetVP}, MobileNet-v2\cite{Sandler2018MobileNetV2IR}. The size of these architectures is specified manually by reducing their blocks, layers in blocks, channels in layers, or their combinations. All the student architecture are specified with $\sim90M$ FLOPs or $\sim0.8M$ parameters.

The performance of different students is shown in Figure \ref{fig:perf}. In subfigure \ref{fig:perf_para}, the two plots correspond to two datasets (i.e., CIFAR10 and CIFAR100), as given in titles of plots. In each plot, there are students with SOTA architectures, three manually specified students,  and our student found by our search approach. For each student corresponding to a single color, there are two marks where the circle mark means the performance when trained from scratch, and the cross mark means the performance under distillation. The x-axis describes the number of parameters, which determines the model size, and the y-axis shows the test error on the test dataset. 

We can observe that all cross marks are located below the circle marks of the same color, which means the model trained under distillation outperforms the same one trained from scratch. The students did benefit from the dark knowledge output by teacher model. For different student architectures, the distillation performance (cross marks) could vary considerably, even with a similar number of parameters or FLOPs. Hence, it makes sense to search for a good student architecture to learn the dark knowledge of the teacher.

The student located in the bottom left corner means it shows better performance with fewer parameters. The closer to the origin of coordinates the mark is located, the better the performance of the corresponding student is. Our student outperforms all other students since it is located in the most bottom left position. Besides the model size, the cost of forwarding inferences of student architectures is also important, especially for latency-critical applications. We also compare our student with other baselines in terms of FLOPs. As shown in subfigure \ref{fig:perf_flops}, our student with the least parameters outperforms most of the other students.

Given the teacher trained on CIFAR100 in subfigure \ref{fig:perf_flops}, the student with an architecture similar to Shufflenet-v2 or MobileNet-v2 performs better than the manually specified students, also better than our student. However, they behave much worse under the teacher model trained on CIFAR10. We can conclude that no student architecture can win under all the datasets. The student with SOTA is not the best for all teachers and datasets. The conclusion further confirms the significance of our contribution to search for a good student architecture for a given teacher and a given dataset. 

The contribution of this work is orthogonal to other Knowledge Distillation techniques. Instead of proposing a new method that outperforms all state-of-the-art Knowledge Distillation techniques, we aim to search for better student architectures than the one specified manually. For other Knowledge Distillation techniques, we can similarly search for good student architectures with the loss functions of other Knowledge Distillation techniques. A large number of Knowledge Distillation techniques \cite{Romero2015FitNetsHF,Zagoruyko2017PayingMA,Czarnecki2017SobolevTF} have been proposed since the work \cite{hinton2015distilling} was published. Hence, we only demonstrate the effectiveness of our search method with the most popular KD technique \cite{hinton2015distilling}.

\begin{figure}[h]
    \centering
     \includegraphics[scale=0.45]{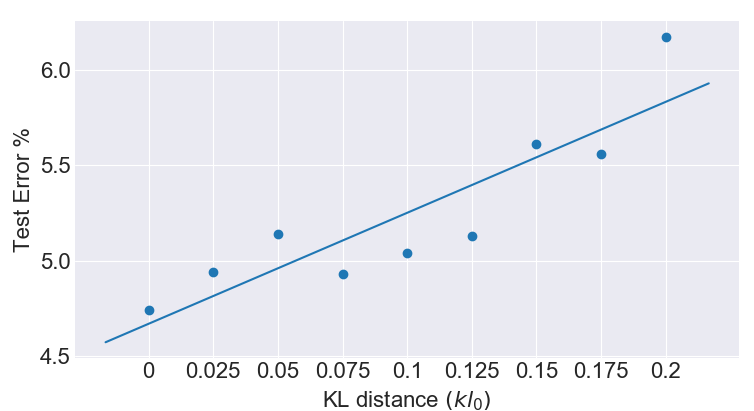}
     \caption{The correlation between KL-divergence and the distillation performance of the selected student}
     \label{fig:corre}
     \vspace{-0.5cm}
\end{figure}

\subsection{Ablation Study and Sensitivity Analysis}
\subsubsection{Ablation Study of the Loss function}
The first term of our loss function describes KL-divergence between the softened outputs of the teacher model and the constructed model. We claim that the term is important to select a good student architecture. To verify this argument, we replace this term in the loss function of Equation \ref{equ:loss} with cross-entropy distance between the normal (not softened) outputs of the teacher model and the constructed model. Then the loss can be treated as a network pruning loss function. Besides, we weight the scaling factors by the normalized number of reduced FLOPs it corresponds to. For comparison, we also conduct experiments by setting all the weights $ \alpha = 1$.

\begin{figure*}[t]
         \centering
         \includegraphics[width=16cm,height=4.5cm]{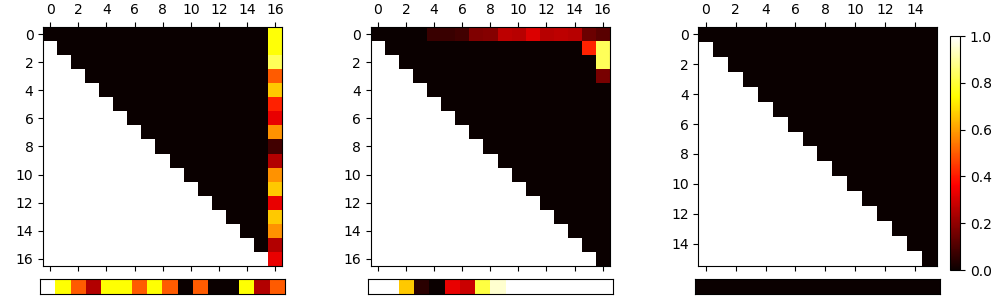}
        \caption{Visualization of the student architecture: each color block corresponds to a skip connection in a denseblock; the color itself indicate its drop ratio of channels in each connection.}
        \label{fig:vis}
\end{figure*}

The experiment results are shown in table \ref{tab:ablation} where NP is Network Pruning, KD is Knowledge Distillation, NOKD is no KD, C10 is CIFAR10 and C100 is CIFAR100. The last row corresponds to the proposed loss. For the first two rows with NP, it prunes the teacher network with the corresponding network pruning loss function and trains the obtained small architecture from scratch or under distillation. By comparing the second row NP$ _{weighted}$ and the last row KD$ _{weighted}$, we can find that the first term of Equation \ref{equ:loss} is important. In other words, the student found by our distillation-aware loss function outperforms the one found by the distillation-agnostic network pruning loss function. The performance corresponding to the loss functions with weighted scaling factors outperforms the one without weighting. We conclude that it is necessary to weight the scaling factors to obtain good student architectures. To be noted that the students found by our loss function perform better even when they are trained without distilled knowledge. It is because the search process has already leveraged the knowledge of the teacher model since the first term in our loss function distills knowledge from the teacher.

\subsubsection{KL-divergence in the Loss Function}
In our loss function, the KL-divergence between the output distribution of the constructed model and that of the teacher model describes how well the selected student learns from the teacher. In the case of KL-divergence = 0, the student has the same generalization ability as its teacher. In the search process, we aim to update the constructed architecture to make the KL-divergence term as close to 0 as possible. We argue that it describes how good the selected student architecture is. We verify our argument using the following loss function.
\begin{equation}
\small
\min_{\mathbf{w, g}}  \frac{1}{N} \sum^N_{i=1} abs(KL(f_s(\mathbf{x_i, w, g}), f_t(\mathbf{x_i})) - kl_0) + R_1 + R_2
\end{equation}
where $R_1$ and $R_2$ are the same as the two regularization terms in Equation \ref{equ:loss}. This loss function differs from our proposed function in the first term. In this new loss function, the optimization pushes the KL-divergence above to $kl_0$ instead of zero. Since the first term in our loss function define what good student architecture is, the $kl_0$ is supposed to correlate with the distillation performance of the selected student architecture. We set $kl_0$ to different values and optimize the constructed model to obtain the student with the FLOPs of $90M$ by early stopping. The maximum of $kl_0$ is the biggest KL-divergence occurring in the training process. The search is conducted under different values of $kl_0$ varying from 0 to the maximum uniformly. 

The selected student architectures under each value of $kl_0$ are trained with knowledge distillation. The results are shown in Figure \ref{fig:corre}. The correlation between test error and $kl_0$ is 0.8896, which means this KL-divergence term in the loss function does define what a good student architecture should be.

\begin{figure}[t]
    \centering
    \includegraphics[width=8.5cm,height=4.5cm]{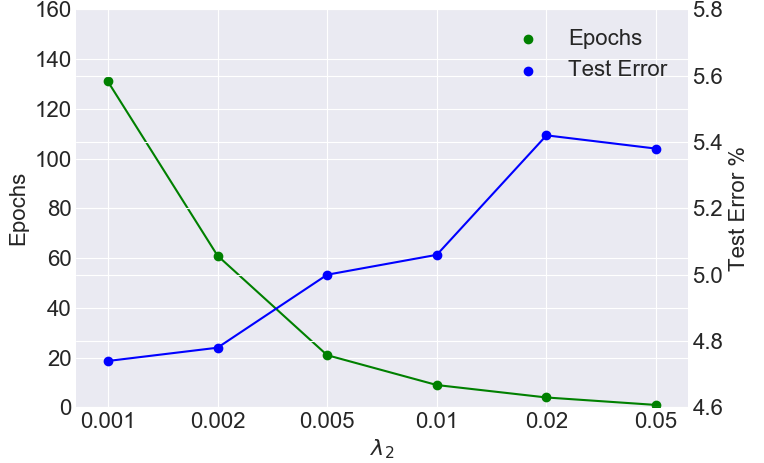}
    \caption{Under different values of $\lambda_2$, the left y-axis shows the number of epochs that are required to find the desired student and the right y-axis shows the test errors of the selected architectures trained on the training data under distillation.}
    \label{fig:relation}
\end{figure}

\subsubsection{Sensitivity Analysis of Hyperparameters}
In the proposed loss function in Equation \ref{equ:loss}, the second term is the widely-recognized weight decay regularization technique. The hyperparameter $\lambda_1$ is set the same in all the training settings. The third term therein works by regularizing the scaling factors and pushing them to zeros. The hyperparameter $\lambda_2$ decides the speed to reach an architecture with the pre-defined number of FLOPs. The baseline and our found architecture are compared under FLOPs of 90M. Concretely, we stop the optimization if the search process finds the student architecture with the pre-defined number of FLOPs. We apply different values of $\lambda_2$ and record the epoch when we find the desired student architecture as well as the final distillation performance of selected students. The relationship among $\lambda_2$, the epoch, and the test error is shown in Figure \ref{fig:relation}. The high value of $\lambda_2$ leads to a quick update to the desired student architecture, while a small $\lambda_2$ requires more epochs to find a desired student architecture. Search with more epochs (i.e., small $\lambda_2$) can select better students.

\subsection{Visualizing and Understanding the Student}
The pioneering work \cite{zeiler2014visualizing} understands the classification decision by visualizing the saliency maps using deconvolutional operations. Recent work \cite{Selvaraju2017GradCAMVE,Gu2018UnderstandingID} creates more meaningful saliency maps to explain the individual classification decisions. Different from those work, we visualize the architecture instead of creating explanations for individual classification decisions. Although the numerical evidence shows that the student found by our search approach is better, it is desirable to explore what the selected student architecture looks like.

The teacher architecture has three dense blocks, and a few channels are removed to obtain a small student architecture. After each dense block, there is also a transition block containing a single convolutional layer and a pooling layer to reduce the size of feature maps. The student architecture is visualized in Figure \ref{fig:vis}. The three plots correspond to the 1st, the 2nd, the 3rd dense block, respectively. The bar under each plot corresponds to the single convolutional layer in the transition block after each dense block. In a sing plot, each small color-block corresponds to a connection. E.g. in 8th layer, there are one stem connection and 8 skip connections, which represented by 9 color blocks in the 8th column. The value in color map varies from 0 to 1, which describe the drop rate of the corresponding connection (the number of removed channels divided by the number of channels before removal).  Drop rate = 0 (white color) means that the corresponding connection is removed.

In third dense block, almost no channel is removed. This observation coincides with the common rule in network engineering, which requires the later stage of a neural network to have more expressive power since more complicated patterns may emerge as the receptive field increases. In the column of the 16th layer, no connection (only channel) is removed although a few channels are removed. The discovery indicates that feature reuse is important to keep the DenseNet powerful, which is inconsistent with claims in the paper \cite{Huang2017DenselyCC}. The removed channels are the ones whose removal can save much computational cost in forwarding inference, which means weighting scaling factors does have an impact on the selection of architectures.

\section{Conclusion}
In this work, we propose to search for a student architecture to learn knowledge from teacher model. The student architecture found by our search process outperforms the ones by manually reducing the size of teacher models as well as the ones with SOTA architectures. We also found interesting experimental results. A found student architecture also performs better when trained from scratch without knowledge distillation, which means the found architecture encodes knowledge of teacher during the search process. How to encode more prior knowledge into architectures can be further explored in future work.

\bibliography{ecai}

\begin{thebibliography}{10}

\bibitem{Ba2014DoDN}
Jimmy Ba and Rich Caruana, `Do deep nets really need to be deep?', in {\em
  NeurIPS}, (2014).

\bibitem{bucilua2006model}
Cristian Buciluǎ, Rich Caruana, and Alexandru Niculescu-Mizil, `Model
  compression', in {\em Proceedings of the 12th ACM SIGKDD}, pp. 535--541. ACM,
  (2006).

\bibitem{Courbariaux2016BinarizedNN}
Matthieu Courbariaux, Itay Hubara, Daniel Soudry, Ran El-Yaniv, and Yoshua
  Bengio, `Binarized neural networks: Training deep neural networks with
  weights and activations constrained to +1 or -1', in {\em NeurIPS}, (2016).

\bibitem{Crowley2018MoonshineDW}
Elliot Crowley, Gavin Gray, and Amos~J. Storkey, `Moonshine: Distilling with
  cheap convolutions', in {\em NeurIPS}, (2018).

\bibitem{Czarnecki2017SobolevTF}
Wojciech Czarnecki, Simon Osindero, Max Jaderberg, Grzegorz Swirszcz, and
  Razvan Pascanu, `Sobolev training for neural networks', in {\em NeurIPS},
  (2017).

\bibitem{Denton2014ExploitingLS}
Emily~L. Denton, Wojciech Zaremba, Joan Bruna, Yann LeCun, and Rob Fergus,
  `Exploiting linear structure within convolutional networks for efficient
  evaluation', in {\em NeurIPS}, (2014).

\bibitem{Gu2018UnderstandingID}
Jindong Gu, Yinchong Yang, and Volker Tresp, `Understanding individual
  decisions of cnns via contrastive backpropagation', in {\em ACCV}, (2018).

\bibitem{han2015learning}
Song Han, Jeff Pool, John Tran, and William Dally, `Learning both weights and
  connections for efficient neural network', in {\em NeurIPS}, pp. 1135--1143,
  (2015).

\bibitem{Han2015LearningBW}
Song Han, Jeff Pool, John Tran, and William~J. Dally, `Learning both weights
  and connections for efficient neural networks', in {\em NeurIPS}, (2015).

\bibitem{hassibi1993second}
Babak Hassibi and David~G Stork, `Second order derivatives for network pruning:
  Optimal brain surgeon', in {\em NeurIPS}, (1993).

\bibitem{he2016deep}
Kaiming He, Xiangyu Zhang, Shaoqing Ren, and Jian Sun, `Deep residual learning
  for image recognition', in {\em CVPR}, pp. 770--778, (2016).

\bibitem{heo2018knowledge}
Byeongho Heo, Minsik Lee, Sangdoo Yun, and Jin~Young Choi, `Knowledge
  distillation with adversarial samples supporting decision boundary', {\em
  AAAI}, (2019).

\bibitem{hinton2015distilling}
Geoffrey Hinton, Oriol Vinyals, and Jeff Dean, `Distilling the knowledge in a
  neural network', {\em stat}, {\bf 1050}, ~9, (2015).

\bibitem{Huang2017DenselyCC}
Gao Huang, Zhuang Liu, and Kilian~Q. Weinberger, `Densely connected
  convolutional networks', {\em CVPR},  2261--2269, (2017).

\bibitem{huang2018data}
Zehao Huang and Naiyan Wang, `Data-driven sparse structure selection for deep
  neural networks', in {\em ECCV}, pp. 304--320, (2018).

\bibitem{Ioffe2015BatchNA}
Sergey Ioffe and Christian Szegedy, `Batch normalization: Accelerating deep
  network training by reducing internal covariate shift', in {\em ICML},
  (2015).

\bibitem{Jaderberg2014SpeedingUC}
Max Jaderberg, Andrea Vedaldi, and Andrew Zisserman, `Speeding up convolutional
  neural networks with low rank expansions', (2014).

\bibitem{kandasamy2018neural}
Kirthevasan Kandasamy, Willie Neiswanger, Jeff Schneider, Barnabas Poczos, and
  Eric~P Xing, `Neural architecture search with bayesian optimisation and
  optimal transport', in {\em NeurIPS}, (2018).

\bibitem{lecun2015deep}
Yann LeCun, Yoshua Bengio, and Geoffrey Hinton, `Deep learning', {\em nature},
  {\bf 521}(7553),  436, (2015).

\bibitem{LeCun1989OptimalBD}
Yann LeCun, John~S. Denker, and Sara~A. Solla, `Optimal brain damage', in {\em
  NeurIPS}, (1989).

\bibitem{lecun1990optimal}
Yann LeCun, John~S Denker, and Sara~A Solla, `Optimal brain damage', in {\em
  NIPS}, pp. 598--605, (1990).

\bibitem{Li2017PruningFF}
Hao Li, Asim Kadav, Igor Durdanovic, Hanan Samet, and Hans~Peter Graf, `Pruning
  filters for efficient convnets', in {\em ICLR}, (2017).

\bibitem{Liu2018HierarchicalRF}
Hanxiao Liu, Karen Simonyan, Oriol Vinyals, Chrisantha Fernando, and Koray
  Kavukcuoglu, `Hierarchical representations for efficient architecture
  search', in {\em ICLR}, (2018).

\bibitem{Liu2019DARTSDA}
Hanxiao Liu, Karen Simonyan, and Yiming Yang, `Darts: Differentiable
  architecture search', in {\em ICLR}, (2019).

\bibitem{Liu2019RethinkingTV}
Zhuang Liu, Mingjie Sun, Tinghui Zhou, Gao Huang, and Trevor Darrell,
  `Rethinking the value of network pruning', in {\em ICLR}, (2019).

\bibitem{Ma2018ShuffleNetVP}
Ningning Ma, Xiangyu Zhang, Hai-Tao Zheng, and Jian Sun, `Shufflenet v2:
  Practical guidelines for efficient cnn architecture design', in {\em ECCV},
  (2018).

\bibitem{Molchanov2017PruningCN}
Pavlo Molchanov, Stephen Tyree, Tero Karras, Timo Aila, and Jan Kautz, `Pruning
  convolutional neural networks for resource efficient inference', in {\em
  ICLR}, (2017).

\bibitem{parikh2014proximal}
Neal Parikh, Stephen Boyd, et~al., `Proximal algorithms', {\em Foundations and
  Trends{\textregistered} in Optimization}, {\bf 1}(3),  127--239, (2014).

\bibitem{Rastegari2016XNORNetIC}
Mohammad Rastegari, Vicente Ordonez, Joseph Redmon, and Ali Farhadi, `Xnor-net:
  Imagenet classification using binary convolutional neural networks', in {\em
  ECCV}, (2016).

\bibitem{Romero2015FitNetsHF}
Adriana Romero, Nicolas Ballas, Samira~Ebrahimi Kahou, Antoine Chassang, Carlo
  Gatta, and Yoshua Bengio, `Fitnets: Hints for thin deep nets', in {\em ICLR},
  (2015).

\bibitem{Sandler2018MobileNetV2IR}
Mark~B. Sandler, Andrew~G. Howard, Menglong Zhu, Andrey Zhmoginov, and
  Liang-Chieh Chen, `Mobilenetv2: Inverted residuals and linear bottlenecks',
  {\em CVPR},  4510--4520, (2018).

\bibitem{Selvaraju2017GradCAMVE}
Ramprasaath~R. Selvaraju, Michael Cogswell, Abhishek Das, Ramakrishna Vedantam,
  Devi Parikh, and Dhruv Batra, `Grad-cam: Visual explanations from deep
  networks via gradient-based localization', {\em ICCV},  618--626, (2017).

\bibitem{Simonyan2015VeryDC}
Karen Simonyan and Andrew Zisserman, `Very deep convolutional networks for
  large-scale image recognition', in {\em ICLR}, (2015).

\bibitem{Wu2016QuantizedCN}
Jiaxiang Wu, Cong Leng, Yuhang Wang, Qinghao Hu, and Jian Cheng, `Quantized
  convolutional neural networks for mobile devices', pp. 4820--4828, (2016).

\bibitem{xie2019exploring}
Saining Xie, Alexander Kirillov, Ross Girshick, and Kaiming He, `Exploring
  randomly wired neural networks for image recognition', {\em arXiv preprint
  arXiv:1904.01569}, (2019).

\bibitem{Zagoruyko2017PayingMA}
Sergey Zagoruyko and Nikos Komodakis, `Paying more attention to attention:
  Improving the performance of convolutional neural networks via attention
  transfer', in {\em ICLR}, (2017).

\bibitem{zeiler2014visualizing}
Matthew~D Zeiler and Rob Fergus, `Visualizing and understanding convolutional
  networks', in {\em European conference on computer vision}, pp. 818--833.
  Springer, (2014).

\bibitem{Zhang2015EfficientAA}
Xiangyu Zhang, Jianhua Zou, Xiang Ming, Kaiming He, and Jian Sun, `Efficient
  and accurate approximations of nonlinear convolutional networks', {\em CVPR},
   1984--1992, (2015).

\bibitem{Zoph2017NeuralAS}
Barret Zoph and Quoc~V. Le, `Neural architecture search with reinforcement
  learning', in {\em ICLR}, (2017).

\end{thebibliography}
\end{document}